\documentclass[pdflatex,sn-mathphys-num]{sn-jnl}

\usepackage{graphicx}%
\usepackage{multirow}%
\usepackage{amsmath,amssymb,amsfonts}%
\usepackage{amsthm}%
\usepackage{mathrsfs}%
\usepackage[title]{appendix}%
\usepackage{xcolor}%
\usepackage{array}%
\usepackage{textcomp}%
\usepackage{manyfoot}%
\usepackage{booktabs}%
\usepackage{algorithm}%
\usepackage{algorithmicx}%
\usepackage{algpseudocode}%
\usepackage{listings}%
\usepackage{subcaption}%
\usepackage{caption}%
\usepackage{lmodern}%
\usepackage{ulem}%

\begin{document}

\title[Article Title]{Trans-defense: Transformer-based Denoiser for Adversarial Defense with Spatial-Frequency Domain Representation}


\author{\fnm{Alik} \sur{Pramanick}}\email{p.alik@iitg.ac.in}

\author{\fnm{Mayank} \sur{Bansal}}\email{b.mayank@iitg.ac.in}
\author{\fnm{Utkarsh} \sur{Srivastava}}\email{utkarsh.srivastava@iitg.ac.in}

\author{\fnm{Suklav} \sur{Ghosh}}\email{suklav@iitg.ac.in}
\author{\fnm{Arijit} \sur{Sur}}\email{arijit@iitg.ac.in}

\affil{\orgdiv{Indian Institute of Technology Guwahati}, \orgaddress{\state{Assam}, \country{India}}}



\abstract{In recent times, deep neural networks (DNNs) have been successfully adopted for various applications. Despite their notable achievements, it has become evident that DNNs are vulnerable to sophisticated adversarial attacks, restricting their applications in security-critical systems. In this paper, we present two-phase training methods to tackle the attack: first, training the denoising network, and second, the deep classifier model. We propose a novel denoising strategy that integrates both spatial and frequency domain approaches to defend against adversarial attacks on images. Our analysis reveals that high-frequency components of attacked images are more severely corrupted compared to their lower-frequency counterparts. To address this, we leverage Discrete Wavelet Transform (DWT) for frequency analysis and develop a denoising network that combines spatial image features with wavelets through a transformer layer. Next, we retrain the classifier using the denoised images, which enhances the classifier's robustness against adversarial attacks. Experimental results across the MNIST, CIFAR-10, and Fashion-MNIST datasets reveal that the proposed method remarkably elevates classification accuracy, substantially exceeding the performance by utilizing a denoising network and adversarial training approaches. The code is available at \url{https://github.com/Mayank94/Trans-Defense}.}




\maketitle

\section{Introduction}\label{secI}
The rapid rise of deep learning in recent years has attracted interest from various industries, as they all aim to integrate this technology into their daily operations to maximize profitability. Applications such as autonomous driving, language translation, facial recognition, biomedical image processing, remote sensing, pattern recognition, speech processing, video/image captioning, image enhancement, and visual question answering rely on deep neural networks (DNNs) as their fundamental architecture. This enables them to leverage automated learning and pattern recognition to perform complex tasks without requiring explicit programming. However, in 2013, Szegedy et al. \cite{szegedy2013intriguing} found a significant weakness of DNNs related to image classification. They found that the neural network could confidently misclassify an image when imperceptible adversarial noises were intentionally added, leading to the concept of adversarial examples~\cite{ranga2023log}. This raises concerns about the robustness of advanced deep learning models in critical applications that affect the safety of millions of people, such as self-driving cars, identity recognition, and the medical domain~\cite{sarker2021deep,zhang2025context,lu2024boosting}.

In response to the critical nature of the problem, researchers have been developing defensive mechanisms to counter these perturbations. These efforts can be broadly categorized into two groups. The first group comprises Model-Specific Defense techniques, which aim to strengthen specific models by enhancing their parameters. Examples include Defensive Distillation~\cite{7546524} and Adversarial Training~\cite{DBLP:journals/corr/KurakinGB16a}. While these methods have proven effective against simpler and early-stage attacks, they are susceptible to more advanced attacks. Additionally, they are computationally intensive, often requiring retraining with adversarial examples. The second group of defenses is referred to as Model-Agnostic. These methods involve pre-processing the image before feeding it to the classifier. For instance, techniques like JPEG Compression~\cite{DBLP:journals/corr/DziugaiteGR16}, Random Pixel Deflection~\cite{10.1007/s11042-023-15883-z}, Image Super Resolution~\cite{DBLP:journals/corr/abs-1901-01677}, and HFG Denoising~\cite{DBLP:journals/corr/abs-1712-02976} fall into this category. Model-agnostic methods are typically faster and considered more favorable than model-specific defenses. Most of the existing defense works have been based on convolutional neural networks (CNNs) and work in the spatial domain without considering the frequency domain information. However, there is significant potential for enhanced and fortified defense mechanisms with the inclusion of frequency domain information and transformers-based networks.

In this work, we propose a denoising strategy that operates in both spatial and frequency domains. Our observations indicate that high-frequency components of attacked images are more severely corrupted compared to their low-frequency counterparts. By mitigating the attack effects in the frequency domain, we can enhance the defense of the model. While the Discrete Wavelet Transform (DWT) has been utilized for frequency analysis in various image processing applications, including denoising, but a basic DWT-based denoiser falls short as attacks become increasingly sophisticated. To address this, we extracted features from both the spatial domain and the DWT wavelets using a transformer layer. We then integrated the features from each domain through cross-attention to reconstruct an image free of attacks. The use of transformers helps capture long-range dependencies, thus facilitating improved reconstruction of the clean image. To further increase the robustness of our defense, we applied adversarial retraining to address any vulnerabilities introduced by the attack, thereby enhancing the security of our classifier. 
Our final defense strategy is two-pronged: the denoiser eliminates perturbations from the image, while adversarial retraining fortifies the classifier, making the overall defense more resilient against a wide range of attacks. Key contributions of our model include:
\begin{itemize}
    \item We propose a novel approach to remove adversarial perturbations added to the images by introducing a denoising network as a pre-processing step. This network allows us to restore perturbed images to their natural domain before they are fed into the target classifier.
    \item We utilize cross-attention to combine low and high-frequency information, which integrates complementary data from different frequency bands to enhance the reconstruction of the image.
    \item We extract features at different scales and use cross-attention to fuse these features, ensuring accurate reconstruction of fine and coarse details in the denoised image.
    \item Further, we propose retraining the classifier using the denoised images. Essentially, this retraining involves training with augmented data, allowing the classifier to learn the pseudo-natural domain of images generated by our pre-processing step.

\end{itemize}

\section{Related Work}\label{secRW}

In this section, we will review related works on the defensive methods of resisting adversarial examples. The adversarial defense involves two main approaches: preprocessing-based and adversarial training-based. Image alteration processes involving randomizing, scaling, padding, compression, and decompression are part of pre-processing-based methods. Before classifying the perturbed image, some pre-processing steps are conducted. In order to handle perturbation as high-frequency noise, Zihao et al.~\cite{liu2019feature}'s transformation techniques provided a two-stage approach. The first step filtered the input picture by performing image quantization in the frequency domain. Using a DNN-based quantization approach, the second step eliminates the distortions in the images caused by over-filtering. Nevertheless, this approach loses a tiny amount of information and is ineffective for slight perturbations. 
Comparable defenses utilizing compression are also suggested in~\cite{bhagoji2018enhancing,das2017keeping,dziugaite2016study,guo2018countering}. A different approach for altering distortion trends in perturbed photos is randomized resizing~\cite{xie2017adversarial}. Adaptive noise reduction (ANR) is one example of scalar quantization and image smoothening that was suggested by Liang et al.~\cite{liang2018detecting}. According to Mustafa et al.~\cite{mustafa2019image}, a strong defense may be obtained by employing wavelet transform-based noise reduction combined with super-resolution to convert a perturbed image to a high-dimensional state. A two-stage lighter transformation approach was developed by Qiu et al.~\cite{qiu2021efficient}. The disturbed picture is first purified using a Discrete Cosine Transform (DCT) quantizer. The remaining pixels are then dispersed widely apart, and random pixels are dropped, making it difficult for an adversary to forecast the transformation pattern of the remaining pixels. Furthermore, it lacks dynamic quality, which implies that it simultaneously ruins the image's valuable elements. There isn't yet a criterion for dynamic randomness that can be used to influence only noise components that are deeply ingrained in a picture. Furthermore, these transformation-based protections are not as effective in the presence of large perturbation rates.
Adversarial training, which involves retraining the neural network using adversarial instances to strengthen it from attacks, is the second crucial defense strategy. A training technique was presented by Rouhani et al.~\cite{darvish2018deepfense} that classified an area that belonged to a collection of data points into an exploited or endangered subdomain using the probability density function. The fact that it is dependent upon its capacity to identify adversarial images is a limitation. In order to increase resilience, a pixel elimination prevention strategy is put forth in~\cite{prakash2018deflecting} that changes the count of the disturbed pixels to the benign category. Liang et al.~\cite{liang2018detecting} use spatial quantization and filtering methods to filter disturbances, seeing them as noise. A process that includes deactivating hidden nodes affected by perturbation was devised by Goswami et al.~\cite{goswami2018unravelling}. 
Adversarial training was carried out using PGD-perturbed training instances by Madry et al.~\cite{madry2017towards}. Magnet~\cite{meng2017magnet} used a network of detectors to mimic many commonly occurring instances in order to guard against malicious perturbations. Several individual transducer and detector networks make up the magnet. It could combat against a black box perturbation. Their approach, though, is ineffective when used in large datasets like ImageNet. Samangouei et al.~\cite{samangouei2018defense} used a GAN to eliminate a perturbed image in a different study. The relationship among the high-frequency information processing of Convolutional Neural Networks (CNNs) and their dependability was examined by Wang et al.~\cite{wang2020high}. A lower perturbation is found to be more dangerous than a larger rate of disruption. ComDefend is a compression paradigm proposed by Jia et al.~\cite{jia2019comdefend} to guard against malicious instances. Clear, processed images served as the training set for this reconstruction network. It offered a poor precision of defense. Using high-temperature softmax labels generated by putting a perturbed image through a teacher model as target labels, Defensive Distillation~\cite{papernot2016distillation} is a teacher-student concept-based adversarial learning technique that trains a target classifier model or student model. The cross-entropy loss gradients in~\cite{ross2018improving} were first regularized; then, it was punished if the model's cross-entropy loss changed as a result. DefenseGAN~\cite{samangouei2018defense} leverages the creative potential of generative models to derive noise-free images that are unaffected by adversary interference. The APE-GAN~\cite{shen2017ape} takes a very similar strategy to the Defense-GAN in that it trains a GAN architecture to safeguard from adversarial perturbations on images. However, the APE-GAN obtains its initial understanding of the kind of attack by utilizing an adversarial image as input. An ensemble defense was further formed by combining several defense architectures~\cite{ju2018relative}. The fundamental idea is that when there are several defense architectures, an attacker would encounter computational difficulties while determining the best gradient path. To improve consistency, Pang et al.~\cite{pang2018towards} suggested a training method employing the Reverse Cross-Entropy (RCE) loss function. One drawback is that it took a lot of time since the range where the features were distributed within the high-dimensional manifold went further. A strong defense from adversarial attacks has also been demonstrated by using a Denoising Autoencoder (DAE) to rebuild the perturbed image restored to the high-dimensional manifold by minimizing the perceptual loss among the perturbed and benign images~\cite{hlihor2020evaluating}. 
In order to prevent blurriness in the reconstructed image, the bottleneck must hold enough information, which might reduce efficiency. In~\cite{zhang2021defense}, a defensive mechanism utilizing residual networks was presented to rebuild a picture that had been perturbed. In order to leverage the generated high-frequency input for learning, a densely high-frequency convolutional neural network (DifNet)~\cite{hu2021difnet} employed edge detection across the activation maps of two convolutional layers. They performed well against a variety of perturbations. \\
\textbf{Limitations of existing works.} As per our review of existing works, we can say that Model Agnostic defenses are better than Model based defense because of their plug and play approach. Adversarial Training has proven to be an effective method but it requires the prior knowledge of the attack to be defended against to generate the examples and adversarial example generation is computationally very expensive. To counter this image preprocessing methods have proven to be effective but these method are as good as the preprocessing methods. Some of the methods such as Defense GANs are solving two optimization problems making it very slow to train which is something we want to solve so that we can perform the defense in real time systems. 
All the preprocessing methods are bottle-necked by their ability to find the closest image in the natural domain given the reconstructed image instead of training the classifier to learn this pseudo-natural domain of the purified images leaving some room to further fortify the defense. A good defense is supposed to be the one which is computationally fast to be deployed on real time systems, robust against any type of attack, can defend any model without any specific retraining or knowledge of the model, all these features have not been found in any of the defenses discussed in the literature and we aim to find the sweet spot defense in this work.

\begin{figure}
    \centering
    \includegraphics[width = \textwidth]{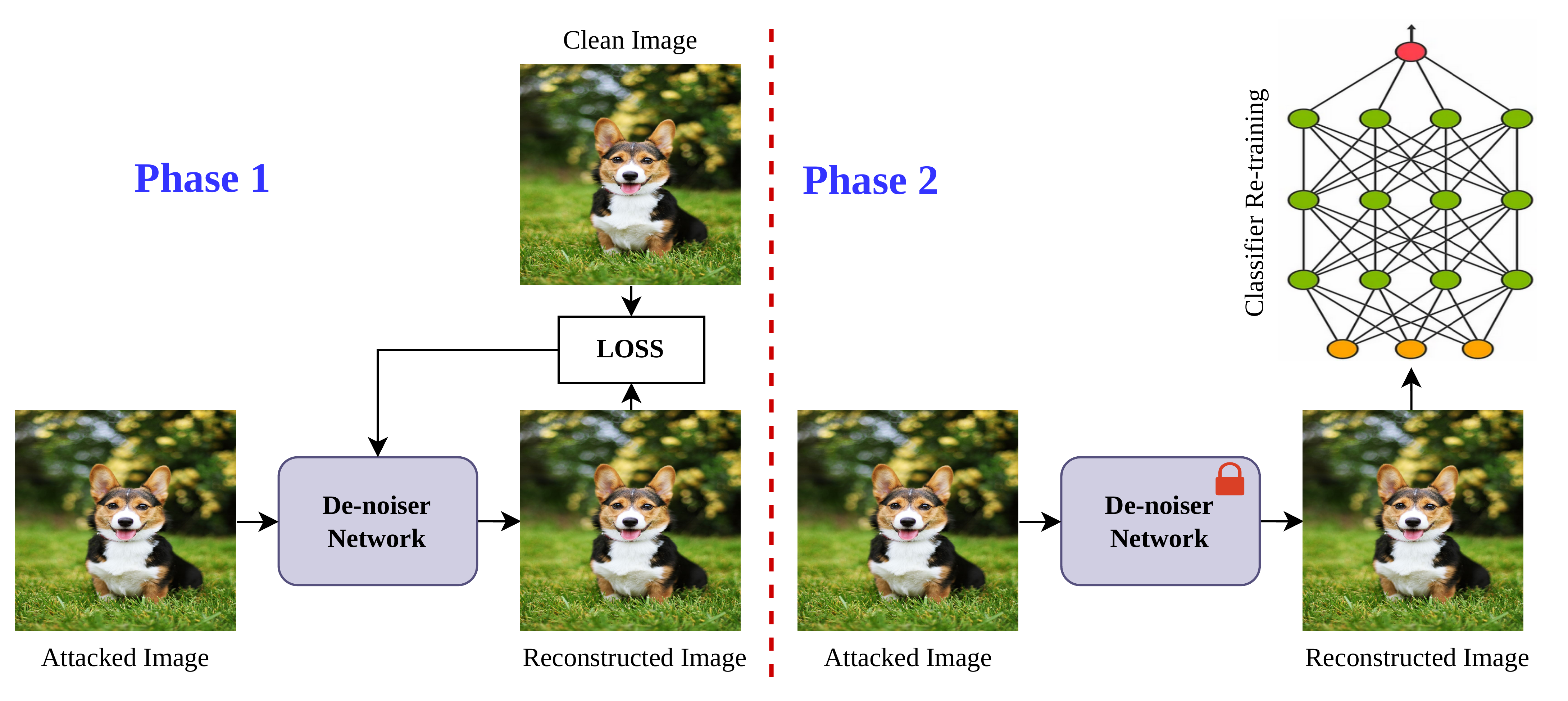}
    \caption{Overview of training pipeline.}
    \label{fig:train}
\end{figure}
\begin{figure}
    \centering
    \includegraphics[width = \textwidth]{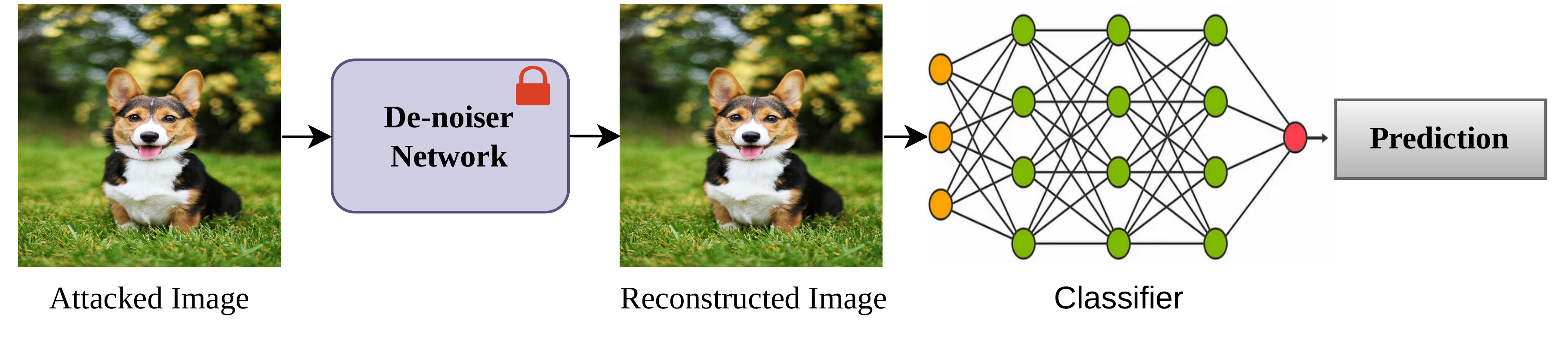}
    \caption{Overview of inference pipeline.}
    \label{fig:infer}
\end{figure}

\section{Proposed Method}\label{secPM}
The workflow of the proposed model is illustrated in Figures \ref{fig:train} and \ref{fig:infer}. Our approach employs a transformer-based denoiser as the core component of our defense strategy. Additionally, we use Discrete Wavelet Transform (DWT) to enhance the denoising process, as adversarial attacks often target higher frequency components. Our defense strategy is further strengthened by integrating adversarial training, which provides a comprehensive and effective safeguard for our model.


\begin{figure}
    \centering
    \includegraphics[width = \textwidth]{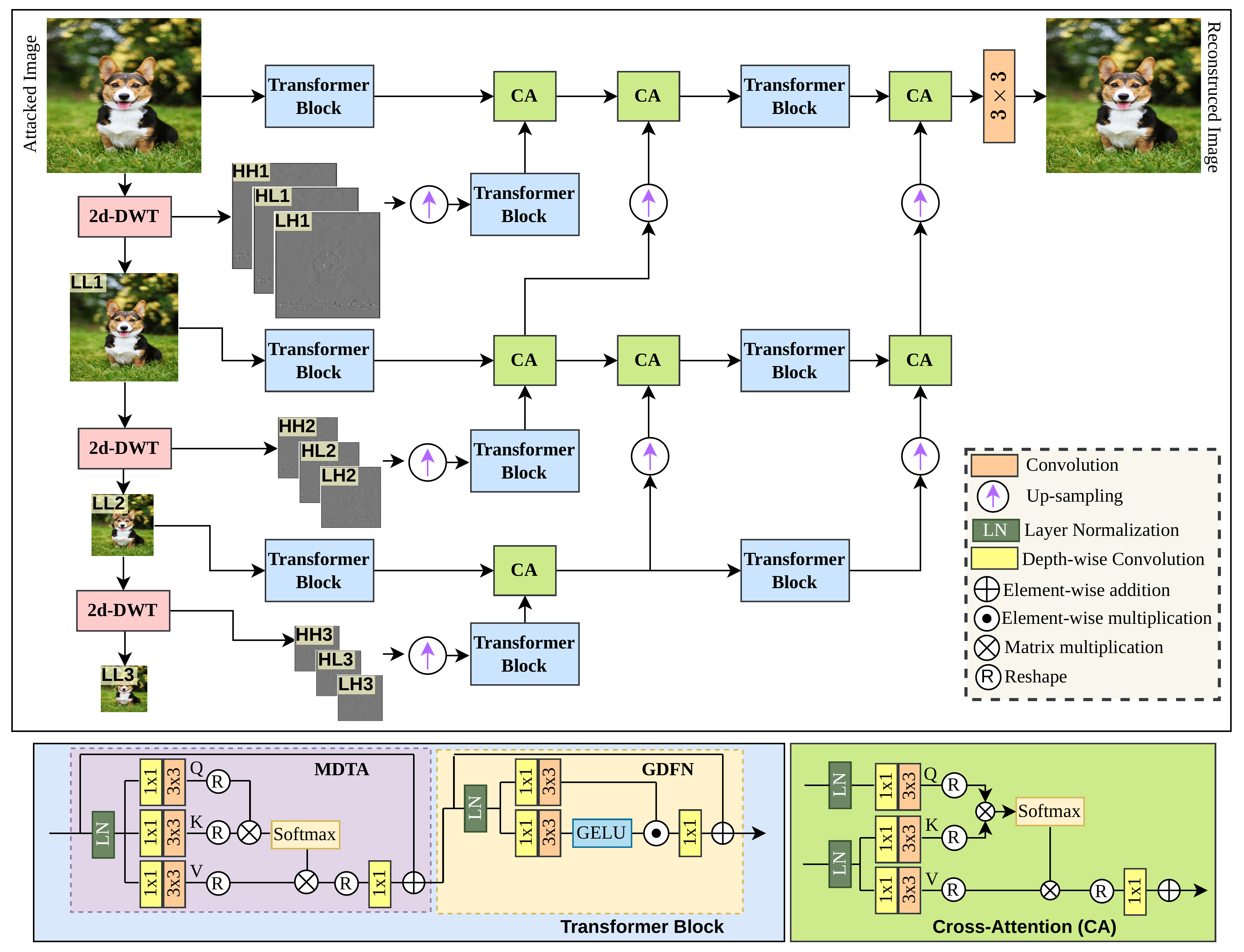}
    \caption{Overview of proposed De-noiser network.}
    \label{fig:my_label}
\end{figure}

\textbf{Overall Pipeline.} We propose a denoiser architecture as illustrated in Figure \ref{fig:my_label}, utilizing Restormer's~\cite{restormer} transformer as a fundamental building block of our model. The architecture inputs both the image and its Discrete Wavelet Transform (DWT) components into the initial layer of transformers to extract features, with cross-attention computed among these extracted feature vectors. These feature vectors are computed at three different scales. To capture information across different scales, cross-attention is also computed among feature vectors from these varying scales. Subsequently, the feature vectors are processed through a second layer of transformer blocks, where additional features are extracted. The final step involves fusing these features and utilizing an output projection layer to reconstruct the purified image.

Let $X_{i}$ be the $ith$ example from the attacked dataset, then we compute DWT wavelets for this sample as
$$ 
X_{LL}, X_{LH}, X_{HL}, X_{HH} = \mathcal{DWT}(X_i)
$$

Now $X_{i}$ is fed to transformer block T1 to extract features say $F_{1}$ and $ X_{LH}, X_{HL}, X_{HH}$ are added, upsampled and then fed to transformer block to get features say $F_{1}^{dwt}$.
\begin{center}
$\begin{aligned} & F_{1} = \mathcal{F}_{Transformer}\left( X_{i} \right) \\ & F_{1}^{dwt} = \mathcal{F}_{Transformer}\left( \left(X_{LH} + X_{HL} + X_{HH}\right) \uparrow_2 \right) \\ & \end{aligned}$
\end{center}

The lower-frequency component of the image ($X_{LL}$), is treated as a standalone image and downscaled by a factor of 2. This approach is feasible because the LL component of the Discrete Wavelet Transform (DWT) contains the spatial features of the image. The process described above is then repeated for this downscaled component.

\begin{center}
$\begin{aligned} & F_{2}=\mathcal{F}_{Transformer}\left( X_{LL} \right) \\ & X^{\prime}_{LL}, X^{\prime}_{LH}, X^{\prime}_{HL}, X^{\prime}_{HH} = \mathcal{DWT}(X_{LL}) \\ & F_{2}^{dwt} = \mathcal{F}_{Transformer}\left( \left(X^{\prime}_{LH} + X^{\prime}_{HL} + X^{\prime}_{HH}\right) \uparrow_2 \right) \\ & \end{aligned}$
\end{center}

This is repeated for one more scale:
\begin{center}
$\begin{aligned} & F_{3}=\mathcal{F}_{Transformer}\left( X^{\prime}_{LL} \right) \\ & X^{\prime \prime}_{LL}, X^{\prime \prime}_{LH}, X^{\prime \prime}_{HL}, X^{\prime \prime}_{HH} = \mathcal{DWT}(X^{\prime}_{LL}) \\ & F_{3}^{dwt} = \mathcal{F}_{Transformer}\left( \left(X^{\prime \prime}_{LH} + X^{\prime \prime}_{HL} + X^{\prime \prime}_{HH}\right) \uparrow_2 \right) \\ & \end{aligned}$
\end{center}
After this, we combined the learning from the features using cross-attention.
\begin{center}
$\begin{aligned} & F_{combined}^{1} = \mathcal{CA}\left( F_{1}, F_{1}^{dwt} \right) \\ & F_{combined}^{2} = \mathcal{CA}\left( F_{2}, F_{2}^{dwt} \right) \\ & F_{combined}^{3} = \mathcal{CA}\left( F_{3}, F_{3}^{dwt} \right) \\ & \end{aligned}$
\end{center}

Finally, we combine these features using cross-attention and reconstruct the denoised image using the output projection layer. Once the denoiser is trained, we freeze the network and perform adversarial training. In this phase \textcolor{red}{2}, we retrain our classifier using images that have been processed through the denoiser network with adversarial attacks. This augmented training process helps address learning gaps and enhances the overall accuracy of the classifier.

\subsection{Transformer Block}
The transformer used in our model must be computationally efficient to approach real-time defense capabilities. Standard transformers are often computationally expensive due to the need for global self-attention across all tokens. To address this, we propose using the Restormer~\cite{restormer} architecture (see Figure~\ref{fig:my_label}), which incorporates key design modifications in the transformer building blocks (multi-headed attention and feed-forward network) to enhance efficiency.

Specifically, the Restormer employs a Multi Dconv head transposed attention (MDTA) block, which has linear complexity, in contrast to the vanilla multi-head self-attention (SA) mechanism. This approach applies self-attention across the feature dimension rather than the spatial dimension while integrating local context through a $1\times1$ convolution before channel-wise aggregation. This combination of pixel-wise and channel-wise aggregation captures both local and global contextual information effectively.

Additionally, the feed-forward network (FFN) of the Restormer is optimized by incorporating a gating mechanism. This mechanism involves an element-wise product of two linear projection layers, with one layer activated using GELU. The gating mechanism regulates which features are complementary and should be propagated forward, enabling subsequent layers to focus on more refined attributes.

Let, $x$ be an intermediate feature. Formally, the operation of the transformer can be defined as:
\begin{equation}
    f_{Transformer} = GDFN(MDTA(x)).
\end{equation}
\subsection{Cross Attention}
Cross-attention integrates two distinct embedding sequences of identical dimensions by employing queries from one sequence and keys and values from the other. Attention masks from one embedding sequence are used to emphasize the extracted features in the other sequence, enabling the combination of learnings from both the spatial and frequency domains for effective image denoising. Specifically, we use cross-attention to merge learnings from low-frequency components ($X_{LL}$) with those from high-frequency components ($ X_{LH}, X_{HL}, X_{HH}$). The query vectors (Q) are derived from the low-frequency components, while the key (K) and value (V) vectors are sourced from the high-frequency components.

Let, A and B are two different embedding sequences then the cross-attention $\mathcal{CA} \left(A, B\right)$ is defined as:
\begin{center}
$\begin{aligned} & Q=\operatorname{DWConv}_{3 \times 3}\left(\operatorname{Conv}_{1 \times 1}\left(A\right)\right), \\ & K, V=\operatorname{DWConv}_{3 \times 3}\left(\operatorname{Conv}_{1 \times 1}\left(B\right)\right) \text {, } \\ & Cross Attention(Q, K, V)= Softmax(Q \cdot K) \cdot V \\ & \mathcal{CA}\left(A,B \right) = {Conv}_{1 \times 1}\left(Cross Attention(Q, K, V)\right)  \\ & \end{aligned}$
\end{center}
\subsection{Discrete Wavelet Transform}
Wavelets represent a time series signal using linear combinations of an orthogonal basis. Depending on the chosen basis, different types of wavelets can be used, such as Haar, Cohen, and Daubechies. The 2D Discrete Wavelet Transform (DWT)~\cite{heil1989continuous} applied to an image  $X_i$ at level 1 yields a low-pass sub-band, known as the approximation coefficients ($LL_i$), as well as high-pass sub-bands, referred to as detail-level coefficients ($LH_i$, $HL_i$, $HH_i$).

Upon examining the wavelet transforms of both attacked and clean datasets, we observed that most perturbations are concentrated in the higher-frequency components (LHLH, HLHL, and HHHH), while the LLLL component is relatively less affected. Although a naive approach would be to remove the detail coefficients, doing so would degrade clean accuracy because model predictions are correlated with these higher-frequency components. Therefore, combining features from the spatial domain with wavelet features using cross-attention results in a more robust denoising network.

\subsection{Loss Function}

\begin{enumerate}
    \item \textbf{Denoiser Loss Function} - We used L1 Charbonnier Loss for training our denoiser which provided a more stable performance than L2 Loss in our use case.
$$
CharbonnierLoss(x_i, y_i) = \frac{1}{N} \sum_{i=1}^{N} \sqrt{(x_i - y_i)^2 + \epsilon^2}
$$
    \item \textbf{Adversarial Training Loss Function} - For Adversarial Training i.e. retraining of the network using trained denoiser, we used Cross Entropy Loss,
    $$
    L(y, \hat{y}) = -\frac{1}{N} \sum_{i}^{N} y_i \log(\hat{y}_i)
    $$
\end{enumerate}

\section{Experiments}\label{secE}
The implementation procedure and results are discussed in this section vividly.

\subsection{Datasets}\label{subsecD}
For this work, three prominent datasets were utilized to assess the efficacy of the proposed methodologies: MNIST~\cite{deng2012mnist}, Fashion-MNIST~\cite{xiao2017fashion}, and CIFAR10~\cite{krizhevsky2009learning}. MNIST and Fashion-MNIST are datasets of grayscale images of handwritten digits and fashion items, respectively, each with 60,000 images of size 28x28 pixels, divided into 50,000 training examples and 10,000 test examples. CIFAR10, on the other hand, consists of 60,000 color images categorized into 10 classes, with 50,000 images for training and 10,000 for testing, each image sized at 32x32 pixels. These datasets were selected to evaluate our methodologies across varying levels of image complexity, from basic grayscale digits and fashion items to more complex color photographs, thereby assessing the generalization and robustness of our approach.

\subsection{Training Details}\label{subsecTD}
We assess the model-agnostic capabilities defense mechanism using two classifiers, ResNet34 and ResNet50. All images were resized to a uniform size of $C \times 32 \times 32$, where $C=1$ for grayscale images (MNIST, FashionMNIST) and $C=3$ for coloured images (CIFAR10). The learning rate ranged from 0.004 to 0.0005, adjusted dynamically over epochs using a scheduler, and Adam optimizer was employed with default beta and epsilon values. The attack intensity during training/testing was fixed as $\epsilon = 0.3/0.2$, $0.3/0.1$, and $0.15/0.05$ for MNIST, FashionMNIST, and CIFAR10, respectively. The denoiser employed had a fixed embedding dimension of 48 and 4 attention heads. All implementations are done on a system with NVIDIA  A-100 GPU, and Pytorch library.

The defense was tested against several adversarial attacks, including the Fast Gradient Sign Method (FGSM), Projected Gradient Descent (PGD) Attack, Momentum Iterative FGSM (MI-FGSM), and Basic Iterative Method (BIM). We first measured the classifiers' accuracy on clean images to establish a baseline. Then, we evaluated their perturbed or attacked accuracy under the influence of each adversarial attack on the images of the same dataset.

We further compared our method against several state-of-the-art defense techniques, including Adversarial Training~\cite{madry2017towards}, Defensive Distillation~\cite{papernot2016distillation}, HFG Denoiser~\cite{li2022feature}, GCTHFA-GAN~\cite{kumar2022globally}, and Collaborative Defense GAN~\cite{laykaviriyakul2023collaborative}.

\subsection{Experimental Results }\label{subsubsecER}
\begin{figure}[!ht]
    \centering
    \begin{subfigure}[b]{\textwidth}
        \centering
        \includegraphics[width=\textwidth]{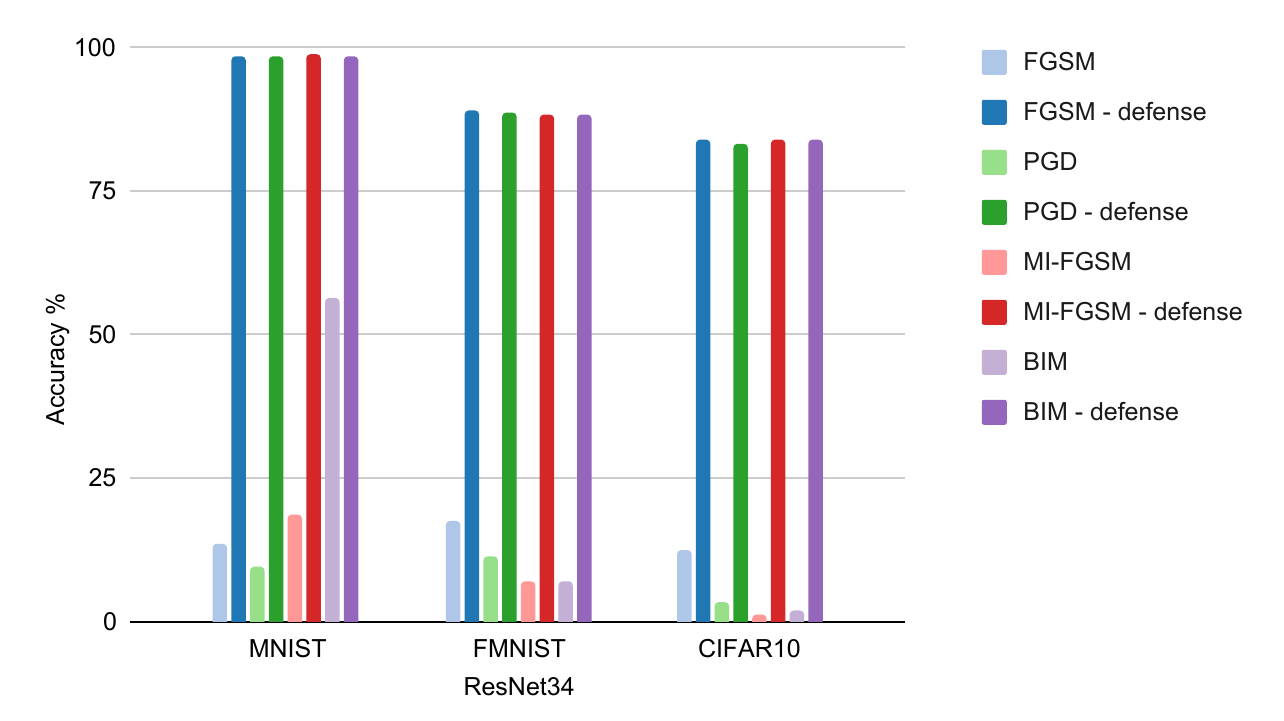}
        \caption{Results on ResNet34 Classifier}
        \label{fig:subfigA}
    \end{subfigure}
    \vspace{0.5cm}
    \begin{subfigure}[b]{\textwidth}
        \centering
        \includegraphics[width=\textwidth]{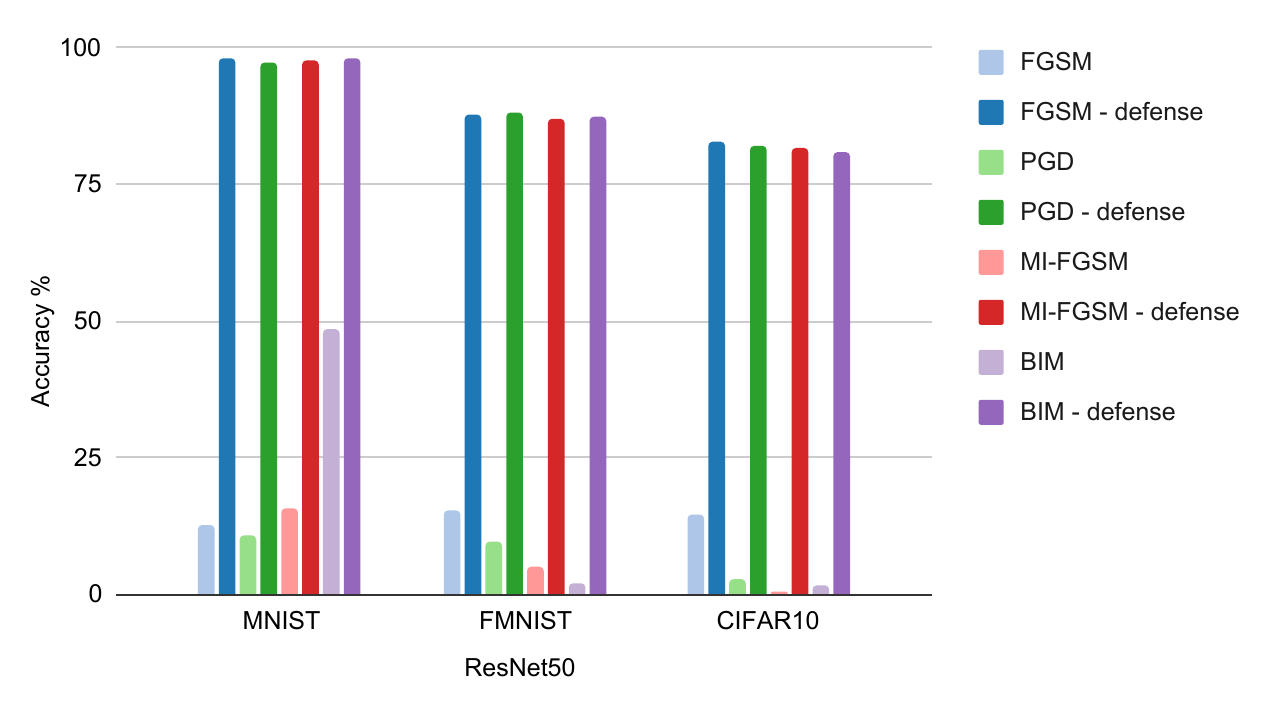}
        \caption{Results on ResNet50 Classifier}
        \label{fig:subfigB}
    \end{subfigure}
    \caption{Comparison of ResNet34 and ResNet50 Classifier Accuracy Against Adversarial Attacks and Our Defense.}
    \label{fig:barGraph}
\end{figure}

The quantitative results for adversarial defense on the target classifiers against various adversarial attacks with varied datasets are described in this section.
The accuracy results of the target classifiers ResNet34 and ResNet50 on MNIST, Fashion-MNIST, and CIFAR10 datasets before and after the adversarial attacks are shown in Figure \ref{fig:subfigA} and \ref{fig:subfigB}.


The classifiers initially achieved near-perfect accuracy on benign datasets under standard training and testing conditions. However, when subjected to multiple adversarial attacks without employing proposed defensive mechanisms, both ResNet34 and ResNet50 exhibited significant decreases in accuracy. For instance, on the MNIST dataset, ResNet34 and ResNet50 achieved accuracies as low as 9.67\% and 10.87\% under PGD adversarial attacks, compared to their benign accuracies of 99.10\% and 98.83\%, respectively. On Fashion-MNIST, prior to BIM adversarial attacks, ResNet34 and ResNet50 achieved accuracies of 91.54\% and 90.84\%, which dropped to 3.099\% and 1.87\% post-attack. Similarly, on CIFAR10, both models saw a significant decline in accuracy, from 92.11\% and 90.11\% to 1.31\% and 0.48\% after MIFGSM attacks.

After the implementation of the proposed defense technique, the models demonstrated considerable resilience against adversarial attacks, maintaining accuracies comparable to their performance prior to being targeted. The defensive strategies effectively minimized the impact of adversarial perturbations on ResNet34 and ResNet50 classifiers across multiple datasets. Specifically, on the MNIST dataset, both models regained accuracies close to their initial benchmarks, achieving robust performance of 98.37\% and 97.27\%, respectively, under PGD adversarial attacks. Similarly, on Fashion-MNIST and CIFAR10, the classifiers exhibited significant improvements in maintaining high accuracy levels post-attack, achieving 88.54\% and 83.95\% for ResNet34 and 87.51\% and 81.59\% for ResNet50, respectively. These results demonstrate the effectiveness of the defensive approach in enhancing the models' resilience to adversarial threats across diverse datasets.


\begin{table}[h]
\caption{Comparison of Existing Adversarial Defense Methods}\label{tab:compare}
\begin{tabular*}{\textwidth}{@{\extracolsep\fill}llcccccc}
\toprule
\multirow{2}{*}{\textbf{Dataset}}&\textbf{    Clean}&\textbf{Type of}&\textbf{FGSM} & \textbf{PGD} & \textbf{MIFGSM} & \textbf{BIM} \\
 & \textbf{Accuracy} & \textbf{Defense} & (\%) & (\%) & (\%) & (\%) \\

\midrule
\textbf{MNIST} & {    99.10\%} & {Adversarial Training~\cite{madry2017towards}} & 88.24 & 82.83 & 91.81 & 90.05 \\
 &  & {Defensive Distillation~\cite{papernot2016distillation}} & {84.75} & {75.29} & {85.45} & {87.24} \\
 &  & {HFG Denoiser~\cite{li2022feature}} & {93.84} & {92.42} & {94.24} &{95.03} \\
 &  & {GCTHFA-GAN~\cite{kumar2022globally}} & {94.48} & {95.74} & {94.19} & {95.69} \\
 &  & {Collaborative GAN~\cite{laykaviriyakul2023collaborative}} & {92.40} & {93.93} & {94.26} & {93.89} \\
 &  & \textbf{Ours} & \textbf{98.65} & \textbf{98.37} & \textbf{98.75} & \textbf{98.58} \\
 
\addlinespace

\textbf{FMNIST} & {    91.54\%} & {Adversarial Training~\cite{madry2017towards}} & 75.73 & 75.77 & 70.78 & 76.02 \\
 &  & {Defensive Distillation~\cite{papernot2016distillation}} & 70.71 & 63.81 & 75.38 & 76.41 \\
 &  & {HFG Denoiser~\cite{li2022feature}} & {79.24} & {81.96} & {80.48} & {80.09} \\
 &  & {GCTHFA-GAN~\cite{kumar2022globally}} & {85.19} & {84.63} & {86.50} & {84.24} \\
 &  & {Collaborative GAN~\cite{laykaviriyakul2023collaborative}} & {82.11} & {80.86} & {77.95} & {81.29} \\
 &  & \textbf{Ours} & \textbf{89.13} & \textbf{88.57} & \textbf{89.29} & \textbf{88.54} \\
\addlinespace

\textbf{CIFAR10} & {    92.11\%} & {Adversarial Training~\cite{madry2017towards}} & {61.71} & {64.81} & {58.75} & {63.25} \\
 &  & {Defensive Distillation~\cite{papernot2016distillation}} & {45.62} & {41.47} & {48.63} & {49.30} \\
 &  & {HFG Denoiser~\cite{li2022feature}} & {74.93} & {75.19} & {74.52} & {74.23} \\
 &  & {GCTHFA-GAN~\cite{kumar2022globally}} & {81.30} & {80.96} & \textbf{84.25} & {81.15} \\
 &  & {Collaborative GAN~\cite{laykaviriyakul2023collaborative}} & {47.58} & {51.73} & {47.97} & {50.15} \\
 &  & \textbf{Ours} & \textbf{84.06} & \textbf{83.29} & {83.95} & \textbf{83.91} \\

\bottomrule
\end{tabular*}
\end{table}

We conducted a comprehensive comparative analysis of our defense methodology against several state-of-the-art defense strategies, including Adversarial Training, Defensive Distillation, HFG Denoiser, GCTHFA-GAN, and Collaborative Defense GAN, as detailed in Table~\ref{tab:compare}. Evaluations performed on the MNIST and Fashion-MNIST datasets revealed that our approach consistently outperformed the state-of-the-art defense methods and achieved superior accuracy in mitigating FGSM, PGD, MI-FGSM, and BIM adversarial attacks. For instance, our method gained 4.41$\%$/4.62$\%$ at FGSM, 2.75$\%$/4.66$\%$ at PGD, 4.76$\%$/3.23$\%$ at MI-FGSM, 3.02$\%$/5.10$\%$ at BIM  improvements in classification accuracy against SOTA on MNIST/FMNIST datasets. Additionally, on the CIFAR-10 dataset, our method demonstrated the highest accuracy across most attack scenarios, although it achieved the second-best performance specifically against the MI-FGSM attack. This comparative assessment demonstrates the effectiveness of our defense methodology across various adversarial contexts.

\section{Ablation Study}\label{secAS}
In this section, we conduct an ablation study to evaluate the effectiveness of different components in enhancing defensive methods across various experimental setups.
\subsection{Effect of Adversarial Retraining with trained denoiser}\label{subsecEAD}
As per our model's pipeline, once we train our denoiser, we retrain our classifier using the denoised images to enhance its robustness. Here,  we evaluate its impact on our defense. Results presented in \textcolor{black}{Table~\ref{tab:ablation1}} indicate a notable 10-12\% increase in accuracy through adversarial retraining.

\begin{table}[h!]      
\caption{Comparison without (w/o) Adversarial Retraining}\label{tab:ablation1}  
\begin{tabular*}{\textwidth}{@{\extracolsep\fill}llcccccc}    
\toprule                                                 
\multirow{2}{*}{\textbf{Dataset}} &{\textbf{    Clean}}& {\textbf{Type of}} &\textbf{FGSM} & \textbf{PGD} & \textbf{MIFGSM} & \textbf{BIM} \\           
 & \textbf{Accuracy} & \textbf{Defense} & (\%) & (\%) & (\%) & (\%) \\   
 \midrule                                               
\textbf{MNIST} & {    99.10\%} & {\textcolor{black}{w/o} Adv Retraining} & 90.24 & 89.83 & 91.48 & 92.05 \\ 
 &  & \textbf{Full Model} & \textbf{98.65} & \textbf{98.37} & \textbf{98.75} & \textbf{98.58} \\            
\addlinespace                                                                                               
\textbf{FMNIST} & {    91.54\%} & {w/o Adv Retraining} & 77.73 & 78.77 & 76.78 & 79.12 \\                   
 &  & \textbf{Full Model} & \textbf{89.13} & \textbf{88.57} & \textbf{89.29} & \textbf{88.54} \\            
\addlinespace                                                                                               
\textbf{CIFAR10} & {    92.11\%} & {w/o Adv Retraining} & {71.71} & {74.81} & {68.75} & {73.25} \\          
 &  & \textbf{Full Model} & \textbf{84.06} & \textbf{83.29} & \textbf{83.95} & \textbf{83.91} \\              
\bottomrule         
\end{tabular*}       
\end{table}         

\subsection{Effect of number of heads in transformer}\label{subsecEHT}
We implemented our model with a single head, which achieved satisfactory accuracy on simpler datasets such as MNIST. However, for more complex datasets like CIFAR10, the denoiser struggled to reconstruct images accurately. Increasing the number of heads to four in the denoiser's transformer blocks enabled better capture of complexity, resulting in improved accuracy. This enhancement was particularly evident in challenging/complex datasets, whereas minimal improvements were observed in simpler ones as shown in Table~\ref{tab:ablation2}.


\begin{table}[h!]
\caption{Comparison with different \textcolor{black}{number of heads}}\label{tab:ablation2}
\begin{tabular*}{\textwidth}{@{\extracolsep\fill}llcccccc}
\toprule
\multirow{2}{*}{\textbf{Dataset}} &{\textbf{    Clean}}& {\textbf{No of}} &\textbf{FGSM} & \textbf{PGD} & \textbf{MIFGSM} & \textbf{BIM} \\
 & \textbf{Accuracy} & \textbf{Head} & (\%) & (\%) & (\%) & (\%) \\

\midrule
\textbf{MNIST} & {    99.10\%} & {1} & 97.92 & 98.03 & 97.48 & 97.35 \\
 &  & {4} & \textbf{98.65} & \textbf{98.37} & \textbf{98.75} & \textbf{98.58} \\
 
\addlinespace

\textbf{FMNIST} & {    91.54\%} & {1} & 82.36 & 81.97 & 80.08 & 81.22 \\
 &  & {4} & \textbf{89.13} & \textbf{88.57} & \textbf{89.29} & \textbf{88.54} \\
\addlinespace

\textbf{CIFAR10} & {    92.11\%} & {1} & {65.71} & {62.61} & {59.95} & {63.58} \\
 &  & {4} & \textbf{84.06} & \textbf{83.29} & \textbf{83.95} & \textbf{83.91} \\

\bottomrule
\end{tabular*}
\end{table}


\section{Conclusion}\label{secC}
In this paper, we proposed a novel transformer-based denoiser architecture for adversarial defense. The denoiser uses transformer blocks with DWT to extract the image's spatial and frequency domain information, which are combined using cross-attention. This is further done at three multiple scales to restore the perturbed image to its natural domain. An adversarial retaining of the classifier was then performed to achieve a more robust defense against a wide range of attacks. Finally, our experimental results show better adversarial defense than existing state-of-the-art methods, and an ablation study was conducted to assess the impact of various configurations within the model.

\section{Competing Interests}
On behalf of all authors, the corresponding author states that there is no conflict of interest.

\section{Funding Information}
Not applicable

\section{Author Contribution}
Alik Pramanick and  Mayank Bansal: Methodology, experimentation, and manuscript preparation; Utkarsh Srivastava and Suklav Ghosh: Manuscript preparation; Arijit Sur: Supervision and manuscript preparation.

\section{Data Availability Statement}
Publicly available MNIST, Fashion-MNIST, and CIFAR-10 dataset.

\section{Research Involving Human and/or Animals}
Not applicable

\section{Informed Consent}
Not applicable
\bibliography{sn-bibliography}

\end{document}